\documentclass[sigconf]{acmart}

\AtBeginDocument{%
  \providecommand\BibTeX{{%
    \normalfont B\kern-0.5em{\scshape i\kern-0.25em b}\kern-0.8em\TeX}}}

\setcopyright{acmcopyright}
\copyrightyear{2018}
\acmYear{2018}
\acmDOI{10.1145/1122445.1122456}

\acmConference[Woodstock '18]{Woodstock '18: ACM Symposium on Neural
  Gaze Detection}{June 03--05, 2018}{Woodstock, NY}
\acmBooktitle{Woodstock '18: ACM Symposium on Neural Gaze Detection,
  June 03--05, 2018, Woodstock, NY}
\acmPrice{15.00}
\acmISBN{978-1-4503-XXXX-X/18/06}

\usepackage{amsmath}

\DeclareMathOperator*{\argmax}{arg\,max}

\usepackage{multirow}
\usepackage{graphicx}

\usepackage{enumitem}
\usepackage{fixltx2e}

\usepackage[ruled,vlined]{algorithm2e}

\usepackage{romannum}

\usepackage{graphicx}

\usepackage{subcaption}
\usepackage{cleveref}

\usepackage{soul}
\DeclareRobustCommand{\hlcyan}[1]{{\sethlcolor{cyan}\hl{#1}}}
\DeclareRobustCommand{\hlyellow}[1]{{\sethlcolor{yellow}\hl{#1}}}

\renewcommand\footnotetextcopyrightpermission[1]{} 
\begin{document}

\title{OneStop QAMaker: Extract Question-Answer Pairs \\from Text in a One-Stop Approach}

\author{Shaobo Cui}
\email{yuanchun.csb@alibaba-inc.com}
\affiliation{%
  \institution{DAMO Academy, Alibaba Group}
}

\author{Xintong Bao}
\email{xintong.bxt@alibaba-inc.com}
\affiliation{%
  \institution{DAMO Academy, Alibaba Group}
}

\author{Xinxing Zu}
\email{patrick.zxx@alibaba-inc.com}
\affiliation{%
  \institution{DAMO Academy, Alibaba Group}
}

\author{Yangyang Guo}
\email{guoyang.eric@gmail.com}
\affiliation{%
  \institution{Shandong University}
}

\author{Zhongzhou Zhao}
\email{zhongzhou.zhaozz@alibaba-inc.com}
\affiliation{%
  \institution{DAMO Academy, Alibaba Group}
}

\author{Ji Zhang}
\email{zj122146@alibaba-inc.com}
\affiliation{%
  \institution{DAMO Academy, Alibaba Group}
}

\author{Haiqing Chen}
\email{haiqing.chenhq@alibaba-inc.com}
\affiliation{%
  \institution{DAMO Academy, Alibaba Group}
}


\begin{abstract}
Large-scale question-answer~(QA) pairs are critical for advancing research areas like machine reading comprehension and question answering. 
To construct QA pairs from documents requires determining how to ask a question and what is the corresponding answer.
Existing methods for QA pair generation usually follow a pipeline approach. Namely, they first choose the most likely candidate answer span and then generate the answer-specific question. 
This pipeline approach, however, is undesired in mining the most appropriate QA pairs from documents since it ignores the connection between question generation and answer extraction, which may lead to incompatible QA pair generation, i.e., the selected answer span is inappropriate for question generation.
However, for human annotators, we take the whole QA pair into account and consider the compatibility between question and answer. 
Inspired by such motivation, instead of the conventional pipeline approach, we propose a model named OneStop generate QA pairs from documents in a one-stop approach. 
Specifically, questions and their corresponding answer span is extracted simultaneously and the process of question generation and answer extraction mutually affect each other.
Additionally, OneStop is much more efficient to be trained and deployed in industrial scenarios since it involves only one model to solve the complex QA generation task. 
We conduct comprehensive experiments on three large-scale machine reading comprehension datasets: SQuAD, NewsQA, and DuReader. The experimental results demonstrate that our OneStop model outperforms the baselines significantly regarding the quality of generated questions, quality of generated question-answer pairs, and model efficiency. 
\end{abstract}





\maketitle
\section{Introduction}
Many tasks in the natural language processing community such as machine reading comprehension and question answering~\cite{hermann2015teaching, rajpurkar2016squad,joshi2017triviaqa} rely heavily on large amounts of human-labeled question-answer~(QA) pairs. 
However, manually annotating QA pairs by human~\cite{bordes2015large,rajpurkar2016squad,joshi2017triviaqa} is both costly and time-consuming. Recently, how to automatically extract QA pairs from documents has attracted increasing attention. 

\begin{figure}[ht!]
    \centering
    \includegraphics[width=0.75\columnwidth]{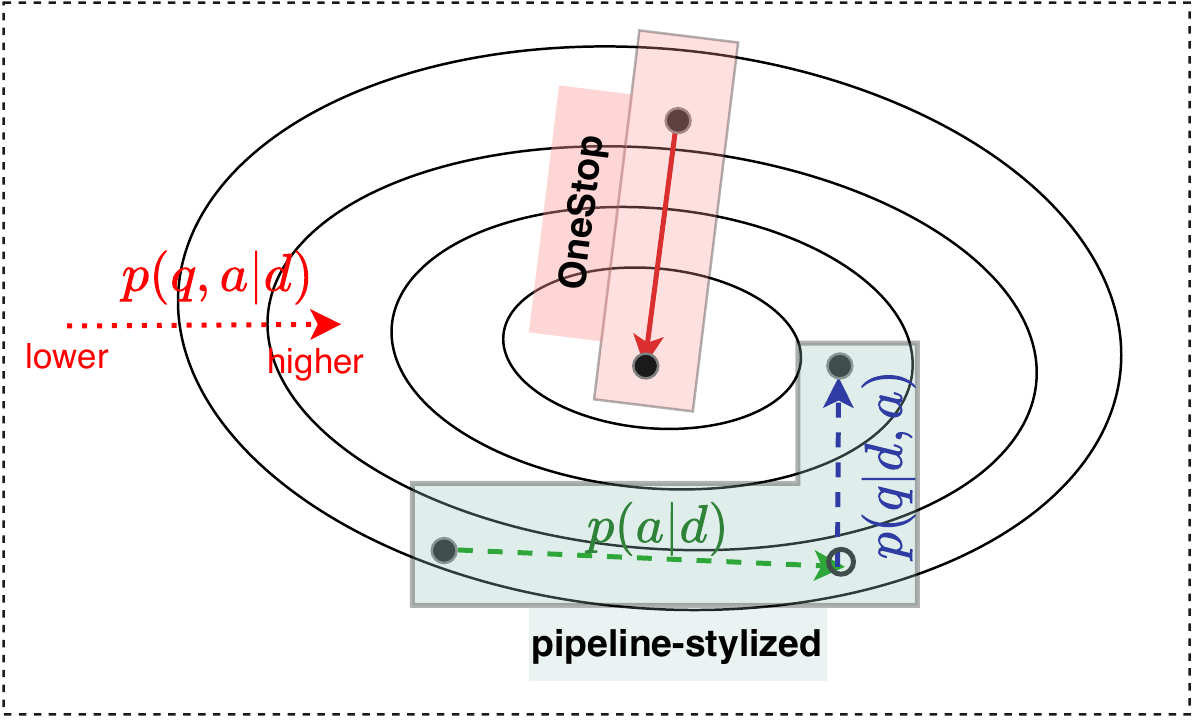}
    \caption{A simplified view of the pipeline approach to QA pair generation. The QA pair under consideration is denoted as a black dot. In the pipeline settings, the expected QA pair is firstly pushed in the direction of maximizing $P(a \vert d)$ and then in the \textit{conditional} direction of maximizing $P(q \vert d, a)$. OneStop approach, however, optimizes in the direction of maximizing $P(q,a \vert d)$ straightforwardly. }
    \label{fig:pipeline_stylized}
\end{figure}

The task of QA pair extraction from a document $d$ is to extract the most related QA pair: $\argmax_{q,a} P(q,a \vert d)$. 
Most of existing works~\cite{yang2017semi,du2018harvesting,alberti2019synthetic,shinoda2020variational,wang2019multi} adopt a pipeline approach, in which firstly selects candidate answer spans from the document: $\argmax_{a} P(a \vert d)$, and then generate the answer-specific questions: $\argmax_{q} P(q \vert d, a)$. 
We present the simplified view of the pipeline approach in Figure~\ref{fig:pipeline_stylized}. 
This type of pipeline approach, however, suffers two major drawbacks. 
Firstly, There is no explicit correlation between the question generation and the answer extraction process. Namely, the question generation model and the answer extraction model are \textbf{isolated} during their training process.  
\begin{table}[hb!]
\caption{Instances to illustrate QA's incompatibility.}
\centering
\resizebox{0.485\textwidth}{!}{
\begin{tabular}{ p{1.3cm}p{1.8cm}p{8.5cm}}
 \toprule
  & Type & Utterance\\
 \midrule
 \multirow{8}{*}{Example 1}   & Approach & First generate a question and then find the generated questions' answer span in document. \\ \cline{2-3}
  & Incompatibility type & Generate a question that is hard to find their answer in document by the answer extraction model. \\ \cline{2-3}
 & \multirow{2}{*}{Document} & The delta is delimited in the West by the in the East by a modern canalized section. \\
  & Question & \hlcyan{What does delta look like?}\\
  & Answer & $--$ \\ 
 \midrule
 \multirow{8}{*}{Example 2}   & Approach & First predict a candidate answer span and then generate an answer-aware question \\ \cline{2-3}
  & Incompatibility type & The predicted answer span is too detailed, incorrect or unsuitable for answer-aware question generation. \\ \cline{2-3}
 & \multirow{2}{*}{Document} & The French crown’s refusal to allow non-Catholics to settle in New France \hlyellow{may help to} explain that colony’s slow rate of population growth compared to that of the neighbouring British colonies, which opened settlement to religious dissenters. \\
  & Answer Span & \hlyellow{may help to}\\
  & Question & $--$ \\
 \bottomrule
\end{tabular}}
\label{tab:incompatible}
\end{table}
This isolation leads the extracted QA pairs to be \textbf{incompatible}: the question generation model may generate questions that are hard to find their corresponding answers by the answer extraction model(see example 1 in Table~\ref{tab:incompatible}), or the answer extraction may extract answer span that are not suitable for question generation(see example 2 in Table~\ref{tab:incompatible}). 
This incompatibility can be explained by Figure~\ref{fig:pipeline_stylized}. These two separate steps: \{ $\argmax_{a}P(a \vert d)$, $\argmax_{q}P(q \vert d,a)$\} are not an accurate approximation for $\argmax_{q,a} P(q, a \vert d)$. 
Secondly, this type of pipeline methods are knotty and time-consuming to be trained and deployed in the industrial online application since they involve at least two models and the cumulative error along the pipeline is huge. 

Unlike the aforementioned pipeline approach, human annotators usually take the whole QA pair into consideration and pay close attention to the compatibility between the extracted answer and the generated question. 
More specifically, from human annotators' perspective, a question that is less likely to be answered by referring to the given document should not be generated in the question generation process. 
Similarly, an answer whose corresponding question is inferior or unsuitable for question generation should be given less attention in the answer extraction process. 
In a nutshell, human annotators consider the compatibility and overall quality of QA pairs. 
Inspired by the limitation of existing pipeline methods and the aforementioned motivation for QA's compatibility, we integrate the question generation and the answer extraction into a unified framework to enhance the compatibility of generated question and the extracted answer. 
We propose OneStop, an architecture which can be easily adapted from existing pre-trained language models to extract QA pairs from documents in a OneStop approach. 
The OneStop model takes documents as input and outputs questions $q$ and questions' corresponding answer spans $a$. 
The answer extraction and the question generation module in the OneStop model collaborate together to find the most compatible QA pairs. Specifically, our OneStop model tackles the objective $\argmax P(q, a \vert d)$ directly instead of the decomposed objectives: $\{\argmax P(a \vert d), \argmax P(q \vert d, a)\}$. These two tasks in our OneStop model mutually affect each other:
(1) the answer span extraction task pushes the question generation model to generate more answerable questions since it is hard to extract the answer span of an unanswerable question;
(2) the question generation task could further enhance the answer extraction model by providing the probability of generating a question. Specifically, the answer extraction model places more attention on questions favored by the question generation model, i.e., the question whose $P(q \vert d)$ is large.
Additionally, by combining the question generation model and answer extraction model in one single model, our OneStop model is much lighter than the existing pipeline approach that involves at least two models. 


As for the model structure, OneStop model adopts the conventional transformer-based sequence-to-sequence structure. Our OneStop model can be easily built upon pre-trained model such as BART~\cite{lewis2020bart}, T5~\cite{raffel2020exploring}, ProphetNet~\cite{qi2020prophetnet} and so on. 
The training objective of the OneStop model is to generate a suitable question and predict the right answer span for this question simultaneously. 
To verify the effectiveness of our OneStop model, we conduct experiment on three large-scale datasets: SQuAD~\cite{rajpurkar2016squad}, NewsQA~\cite{trischler2016newsqa}, and DuReader~\cite{he2018dureader}. We compare the involved baselines in terms of the quality of generated questions, the quality of QA pairs, and model efficiency. Experimental results prove that our OneStop model achieves
SOTA performance in a more efficient way. 
The contributions of this paper are summarized as follows: 
\begin{enumerate}[leftmargin=*]
    \item We propose a unified framework in which the answer extraction module and the question generation module could mutually enhance each other. 
    \item To our best knowledge, OneStop is the first transformer-based model for generating more compatible QA pairs from documents in a one-stop approach. 
    \item OneStop can be easily built upon existing pre-trained language models. Compared with previous pipeline approaches, our OneStop model is much more efficient to train and deploy in industrial scenarios and requires much less human effort.  
    \item We conduct comprehensive experiments on three large-scale datasets to evaluate our OneStop model in terms of question generation, QA pair generation and model efficiency. 
\end{enumerate}
\section{Related Work} \label{sec:literature}
\noindent\textbf{Question Generation.\quad}
Question generation~\cite{yuan2017machine,zhao2018paragraph,sun2018answer,subramanian2018neural,pan2019recent,chan2019recurrent,kim2019improving} is a well-studied natural language processing task in literature.  
There are mainly two types of approaches for question generation: template-based and model-based. Methods~\cite{heilman2010good,labutov2015deep} in the first category rely on human efforts to design the template rule and are thus unscalable across datasets. 
In contrast, the model-based methods ~\cite{yuan2017machine,zhao2018paragraph} employ an end-to-end neural network to generate questions, which takes as inputs selected key phrases and documents. 
However, these methods are limited as the questions cannot be generated from documents directly. An additional entity extraction model or a sequence labeling model~\cite{subramanian2018neural,wang2019multi} is required to determine which part of the document is worthy of being asked.
As a result, it is less practical for this kind of methods in question generation due to the following two facts: 
(1) the key phrase extraction model demands addition manual labor and elaborated tuning;
(2) The most question-worthy phrases in a document are difficult to be identified. 

\noindent\textbf{Question-Answer Pair Generation.\quad}
Most of existing works\quad\cite{du2017learning,indurthi2017generating,alberti2019synthetic,liu2020asking,krishna2019generating,lee2020generating} focusing on the QA pair generation follow a pipeline fashion: (1) determine what points in the document should be asked; (2) Learn to ask based on the selected points; (3) Detect the answer span of the question in the document;
\citeauthor{du2018harvesting} firstly detected the question-worthy answer~(which they dubbed as answer span identification) and then generated the answer-aware question. Similarly, \citeauthor{golub2017two} proposed a two-stage SynNet for QA pair generation, which consists of an answer tagging module and a question synthesis module. 
\citeauthor{alberti2019synthetic} proposed to generate QA pairs with models of question generation and answer extraction and then filtered the results with roundtrip consistency.

\noindent\textbf{Joint Models for Question Generation and Question Answering.} \quad There have been studies~\cite{tang2017question,wang2017joint,song2017unified,cui2019dal} focusing on solving question generation and question answering together. 
In these methods, the input and output of question generation and question answering are inverse, which makes them dual tasks. In this way, question generation and question answering are implemented with separate models connected by their duality. However, the training objective of question answering poses an adverse effect on the performance of the question generation model due to the enforcement of dual constraint. 
Our work is different from these work~\cite{song2017unified,tang2017question,wang2017joint,cui2019dal}, which focus on the duality of question generation and question answering. 
Firstly, for QA extraction from documents, there is not an explicit duality between question generation and answer extraction. Consequently, the duality between these two tasks no longer exists. 
Secondly, question generation and answer extraction in OneStop are optimized in a multi-task learning approach.  Namely, they are optimized simultaneously to find a compatible and optimal solution for QA pair generation. 

\section{Problem Definition} \label{sec:definition}
Given a document, the objective of QA pair generation is to find the related QA pairs. 
Mathematically: 
\begin{equation}
    \bar{q}, \bar{a} = \argmax_{q, a} P(q, a \vert d),
\end{equation}
where document $d$ is a sequence of utterances; answer $a$ should be a sub span from the document and question $q$ is an utterance that is closely associated with $a$. 
Based on this formulation, existing methods can be classified into the following two groups:  
 \begin{figure}[ht!]
\subcaptionbox{D2A2Q \label{fig:comparison:D2A2Q}}{    \centering
    \includegraphics[width=0.75\columnwidth]{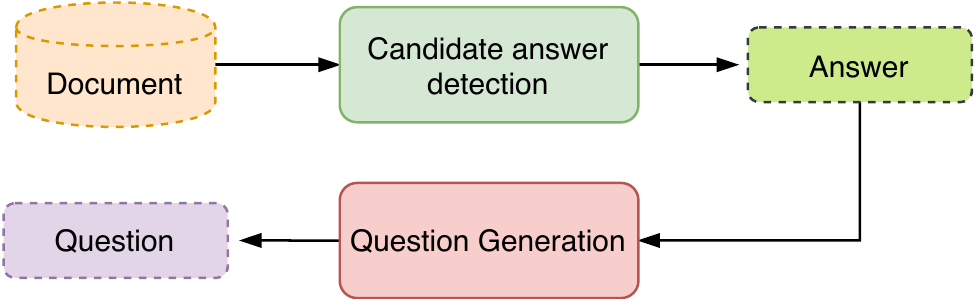}
}
\subcaptionbox{D2Q2A \label{fig:comparison:D2Q2A}}{    \centering
    \includegraphics[width=0.75\columnwidth]{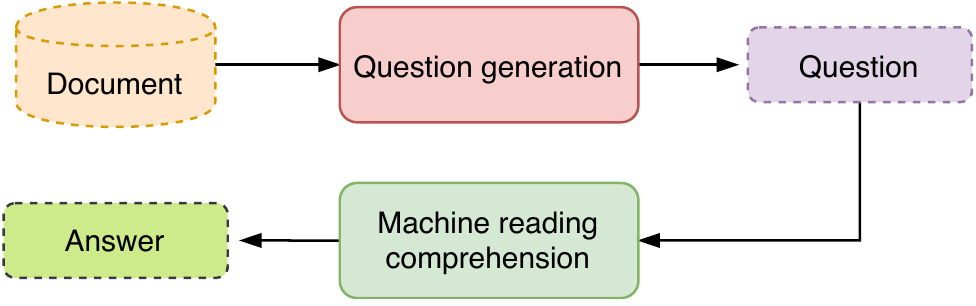}
}
\subcaptionbox{OneStop \label{fig:comparison:OneStop}}{    \centering
    \includegraphics[width=0.75\columnwidth]{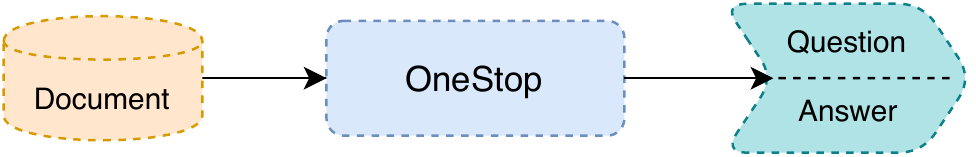}
}
\caption{The comparison of D2A2Q, D2Q2A ,and OneStop.}
\label{fig:comparison}
\end{figure}
\begin{enumerate}[leftmargin=*]
    \item \textbf{D2A2Q}: 
    The candidate answer is first extracted from the document: $P(a \vert d)$, after which the answer-specific question is generated based on the document and the extracted candidate answer: $P(q \vert d, a)$. It can be summarized as: 
    \begin{equation}
         \argmax_{q, a} P(q, a \vert d) \approx 
         \begin{cases}
         \argmax_{a} P(a \vert d; \mathbf{\theta}_{\text{d2a}}), \; & \text{Step \Romannum{1}}  \\
         \argmax_{q} P(q \vert d, a; \mathbf{\theta}_{\text{da2q}}), \; &\text{Step \Romannum{2}}
         \end{cases}
    \end{equation}
    where $\theta_{\text{d2a}}$ and $\theta_{\text{da2q}}$ are the parameters of candidate answer extraction model and answer-specific question generation model respectively.  
    \item \textbf{D2Q2A}: It firstly generates question that is most likely to be asked from the document, i.e., $P(q \vert d; \mathbf{\theta}_{\text{d2q}})$. And then the generated question is utilized to extract its corresponding answer span from the document. Similarly, D2Q2A approach can be summarize as: 
    \begin{equation}
         \argmax_{{q, a}} P(q, a \vert d) \approx 
         \begin{cases}
         \argmax_{q} P(q \vert d; \mathbf{\theta}_{\text{d2q}}) \; & \text{Step \Romannum{1}}  \\
         \argmax_{a} P(a \vert d, q; \mathbf{\theta}_{\text{dq2a}}) \; &\text{Step \Romannum{2}}
         \end{cases}
    \end{equation}
    where $\theta_{\text{d2q}}$ and $\theta_{\text{dq2a}}$ are parameters of question generation model and answer extraction~(machine reading comprehension) model respectively. 
\end{enumerate}

The aforementioned pipeline approaches such as D2Q2A and D2A2Q are all quite rough approximation to the original objective of $\argmax_{q, a} P(q, a \vert d)$~\footnote{For D2Q2A, the final output along the pipeline $\argmax_{q} P(q \vert d)$, $\argmax_{q} P(a \vert d, q)$ are unlikely to be the optimal solution for $\argmax_{q, a} P(q, a \vert d)$. A similar conclusion can be obtained for D2A2Q.}. 
The cumulative error is magnified along these pipelines. Additionally, the training cost and inference efficiency are unfavorable. Motivated by these limits, we propose the OneStop model that models the objective much more precisely. The OneStop framework can be formulated as: 
\begin{equation} \label{eq:onestop_dqa}
\begin{split}
         \argmax_{q, a} P(q, a \vert d) & =          \argmax_{q, a} P(q \vert d; \mathbf{\theta}) \cdot P(a \vert d, q; \mathbf{\theta})  \\
\end{split}
\end{equation}
where $\theta$ is the parameters of the OneStop model. 
The answer extraction module and the question generation module in the OneStop model share the model parameters $\theta$, which means these two tasks influence each other. 
As can be observed, our OneStop model is easier to train and more efficient to use during inference since it involves only one model. 
We present the comparison of these three different approaches in Figure~\ref{fig:comparison}. As we can see, both D2A2Q and D2Q2A are pipeline approaches. Nevertheless, our OneStop model tackle the original QA pair generation objective directly. 

\section{OneStop Model} \label{sec:model}

In this section, we firstly present the overview of the OneStop model in Section~\ref{sec:model:overview}.  Section~\ref{sec:model:qg} and Section~\ref{sec:model:mrc} are about the question generation and answer span extraction module of the OneStop model respectively, after which we end this section with the training and inference of OneStop model in Section~\ref{sec:model:train_and_inference}. 

\subsection{Preliminary: Self-Attentive Module}\label{sec:model:preliminary}
\begin{figure}[H]
    \centering
    \includegraphics[width=0.7\columnwidth]{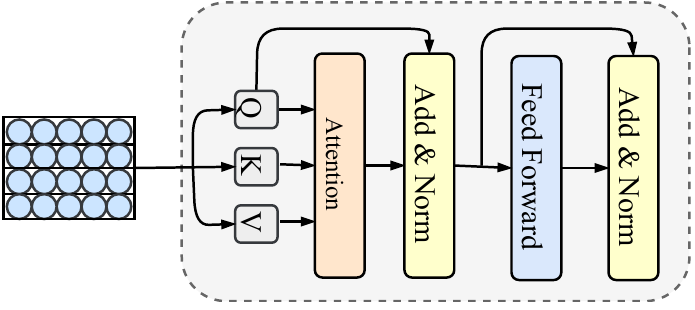}
    \caption{Self-attentive module.}
    \label{fig:self_attentive}
\end{figure}

Inspired by the superiority of transformer~\cite{vaswani2017attention} in utterance representation, we adopt the self-attentive unit as the basic unit of encoder and decoder in our OneStop model. As shown in Figure~\ref{fig:self_attentive}, each self-attentive unit consists of a self-attention layer and a position-wise fully connected feed-forward layer. Each of these two layers is employed with residual connection, followed by layer normalization. 
More specifically, the whole computation process in the self-attentive module can be summarized as: 
\begin{align*}
  \text{Att}(\mathbf{Q}, \mathbf{K}, \mathbf{V}) = \text{Softmax}(\frac{\mathbf{Q}\mathbf{K}^\mathsf{T}}{\sqrt{d_k}})\mathbf{V} \\
  \mathbf{X}_{1} = f_{\text{norm}}(\mathbf{Q}+\text{Att}(\mathbf{Q}, \mathbf{K}, \mathbf{V})) \label{equ:addnorm} \\
  \mathbf{X}_{2} = \max (0, \mathbf{X}_{1} \mathbf{W}_1+\mathbf{b}_1)\mathbf{W}_2+\mathbf{b}_2 \\
  f_{\text{att}}(\mathbf{Q}, \mathbf{K}, \mathbf{V}) = f_{\text{norm}}(\mathbf{X}_1 + \mathbf{X}_{2})
\end{align*}
where $d_k$ is the model dimension.  $f_{\text{att}}(\mathbf{Q}, \mathbf{K}, \mathbf{V})\in \mathbb{R}^{t \times d_k}$, where $t$ is the input length, $\mathbf{Q} \in \mathbb{R}^{t \times d_k}$, $\mathbf{K} \in \mathbb{R}^{t \times d_k}$ and $\mathbf{V} \in \mathbb{R}^{t \times d_k}$.
$\mathbf{W}_1, \mathbf{b}_1, \mathbf{W}_2$ and $\mathbf{b}_2$ are learnable model parameters. 
The computation process of self-attentive module is shown in Figure~\ref{fig:self_attentive}. 

\subsection{Model Overview} \label{sec:model:overview}
\begin{figure}[ht!]
    \centering
    \includegraphics[width=0.95\columnwidth]{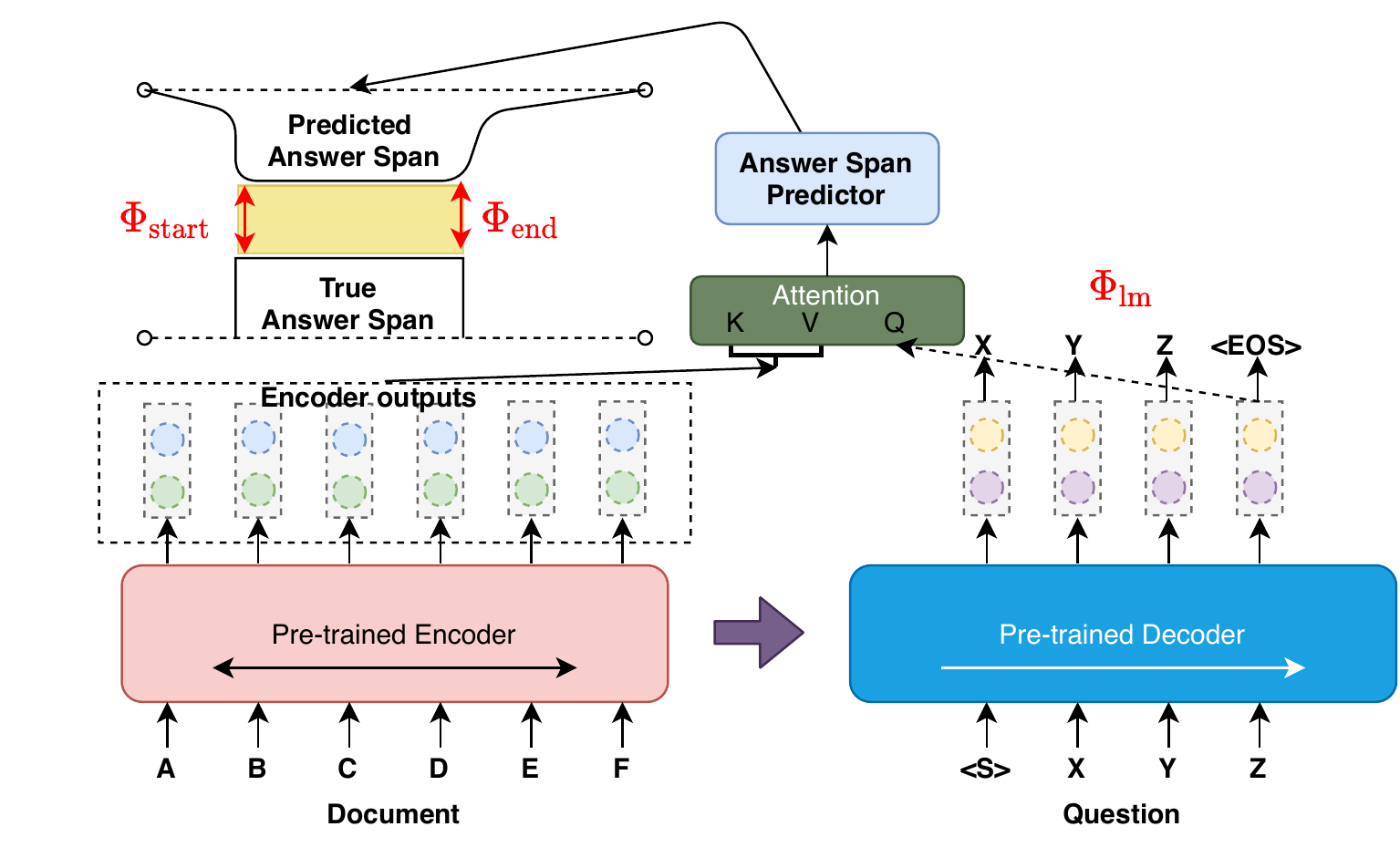}
    \caption{Overview of our proposed OneStop model.}
    \label{fig:model}
\end{figure}

The overview of our OneStop model is presented in Figure~\ref{fig:model}. 
Our OneStop model uses the canonical sequence-to-sequence transformer~\cite{vaswani2017attention} architecture. 
Here we take BART model as an example to illustrate our model structure. Note that our OneStop approach can be easily modified from other pre-trained language models such as T5~\cite{raffel2020exploring} or ProphetNet~\cite{qi2020prophetnet}. 
OneStop model consists of a bidirectional encoder and an auto-regressive decoder. 
The encoder in the OneStop model takes the document as input, and each decoder layer performs cross attention over the final hidden layer of the encoder's outputs. The decoder decodes the generated question in an auto-regressive approach. The start and end position of answer span are predicted based on the encoder outputs and the decoder's outputs at <eos> position. This can be explained by the fact that the corresponding answer for a generated question is determined by document and question together. 

\subsection{Question Generation} \label{sec:model:qg}
As described above, the input of the encoder is the document $d$ and the output of the decoder is expected to be the question $q$. The cross entropy loss for question generation is denoted as: 
\begin{align}
    \Phi_{\text{lm}} = - \sum^{\vert q \vert}_{t=1} \log P(q_t \vert q_{<t}, d; \theta),
\end{align}
where $\vert q \vert$ is the length of the question and $P(q_t \vert q_{<t}, d; \theta)$ is the predicted probability for token $q_t$.  
After we obtain the generated question, we use the decoder outputs at \textrm{<eos>} position as the representation of the generated question, which is denoted as $\mathbf{q}_{\text{eos}}$.
All the outputs of encoders is denoted as: $\mathbf{D}^{e} = \{\mathbf{D}^e_1, \mathbf{D}^{e}_2, \cdots, \mathbf{D}^{e}_{\vert d \vert}\}$,
where $\mathbf{D}^{e} \in \mathbb{R}^{\vert d \vert \times m}$ and $m$ is the model dimension. $\mathbf{D}^e_i$ represents the encoder outputs at position $i$ and $\vert d \vert$ is the length of the input document. 
Compared with RNN-based models like GRU~\cite{cho2014learning} or LSTM~\cite{hochreiter1997long}, the self-attentive module endows us the advantage of encoding each token in a given utterance at the same time. 

\subsection{Answer Span Prediction} \label{sec:model:mrc}
Given the generated question and the document, answer span extraction aims to predict the answer span in the document that could answer the generated question properly. More specifically, it should predict the start and end position of the ground-truth answer span. 
The label prediction network takes $\mathbf{D}^{e}$ and $\mathbf{q}_{\text{eos}}$ as inputs. The probabilities of each token being the start label and the end label are as: 
\begin{equation}
    \begin{cases}
        P_{\text{start}}(i) &= \frac{\exp({\mathbf{D}^e_i \mathbf{W}_s \mathbf{q}_{\text{eos}}})}
        {\sum_{i} \exp({\mathbf{D}^e_i \mathbf{W}_s \mathbf{q}_{\text{eos}}}) } \\
        P_{\text{end}}(i) &= \frac{\exp({\mathbf{D}^e_i \mathbf{W}_e \mathbf{q}_{\text{eos}}})}
        {\sum_{i} \exp({\mathbf{D}^e_i \mathbf{W}_e \mathbf{q}_{\text{eos}}}) }
    \end{cases},
\end{equation}
where $\mathbf{W}_s, \mathbf{W}_e$ are learnable parameters.
The cross entropy loss for the start and the end label prediction are: 
\begin{equation}
    \begin{cases}
        \Phi_{\text{start}}&= - \log P_{\text{start}}(a_{\text{start}} \vert d, q; \theta) \\
        \Phi_{\text{end}}&= - \log P_{\text{end}}(a_{\text{end}} \vert d, q; \theta) 
    \end{cases},
\end{equation}
where $a_{\text{start}}$ and $a_{\text{end}}$ are true positions of the answer's start and end.

\subsection{Training and Inference of OneStop Model} \label{sec:model:train_and_inference} 
\noindent\textbf{Training of OneStop Model\quad} As introduced in Equation~\ref{eq:onestop_dqa}, we have: 
\begin{equation}
\resizebox{0.45\textwidth}{!}{
$
\begin{split}
         P(q, a \vert d; \theta) &= P(q \vert d; \mathbf{\theta}) \cdot P(a \vert d, q; \mathbf{\theta})  \\
         &= \bigg( \prod^{\vert q \vert}_{t=1} P(q_t \vert q_{<t}, d; \theta) \bigg) 
         \cdot \bigg( P_{\text{start}}(a_{\text{start}} \vert d, q; \theta) \cdot P_{\text{end}}(a_{\text{end}} \vert d, q; \theta) \bigg)
\end{split}
$
}
\end{equation}

The negative log-likelihodd of OneStop model can be expressed: 

\begin{equation}
\begin{split}
    \Phi&=-\log  P(q, a \vert d; \theta) \\
    &=-\sum^{\vert q \vert}_{t=1} \log P(q_t \vert q_{<t}, d; \theta) - \log P_{\text{start}}(a_{\text{start}} \vert d, q; \theta) \\
    &\quad-\log P_{\text{end}}(a_{\text{end}} \vert d, q; \theta)\\
    &=\Phi_{\text{lm}} + \Phi_{\text{start}} + \Phi_{\text{end}}
\end{split}
\end{equation}
We use a generalization of OneStop objective which introduces a hyperparameter $\lambda$ that balance question generation and answer extraction: 
\begin{equation} \label{eq:final_loss}
    \Phi = \lambda \cdot \Phi_{\text{lm}} + (1 - \lambda) \cdot(\Phi_{\text{start}} + \Phi_{\text{end}})
\end{equation}

\noindent\textbf{Training Algorithm} \quad
The training algorithm of the OneStop model is described in Algorithm~\ref{algo:training}.

\begin{algorithm}[ht]
\SetAlgoLined
\LinesNumbered
\SetKwInOut{Input}{Input}
\SetKwInOut{Output}{Output}

\Input{$(d, q, a)$ triples, a pre-trained BART language model. }
\Output{OneStop model which takes the document as input and outputs QA pairs.}
Load the pre-trained BART model as the initial checkpoint of generation parts of the OneStop model\;
Fine-tune the question generation part of the OneStop model with $\gamma=1$ in Equation~(\ref{eq:final_loss}), i.e., $\Phi = \Phi_{\text{lm}}$\;
Fine-tune the answer prediction part of the OneStop model with $\gamma=0$ in Equation~(\ref{eq:final_loss}), i.e., $\Phi = \Phi_{\text{start}} + \Phi_{\text{end}}$\;
Determine the value of $\gamma$\;
\While{not converge}{
    Fine-tune the OneStop model with $\Phi= \lambda \cdot \Phi_{\text{lm}} + (1 - \lambda) \cdot(\Phi_{\text{start}} + \Phi_{\text{end}}) $\; 
 }
 \caption{Training algorithm of OneStop model.}
\label{algo:training}
\end{algorithm}

\noindent\textbf{Inference of OneStop Model\quad}
In the inference phase, we feed the document into the OneStop's encoder, the question is generated from OneStop's decoder in an auto-regressive approach. The start and end position of the answer span are predicted by the answer span predictor network. With the start and end position, we can obtain the answer span in the document for the generated question. 
OneStop model also supports for generating multiple QA pairs for long documents, i.e., the long document could be split as multiple sub-documents and OneStop could generate the most related QA pairs for each sub-document.


\section{Experiment Setup}
\label{sec:experiment}
In this section, we mainly elaborate the datasets, involved baselines, evaluation metrics, and model settings sequentially. 
\subsection{Datasets}

In this paper, we conducted experiments on three large-scale machine reading comprehension datasets to evaluate the performance of our proposed OneStop model. 
\begin{itemize} [leftmargin=*]
    \item \textsf{SQuAD}~\cite{rajpurkar2016squad}: SQuAD consists of questions posed by crowdworkers on Wikipedia articles, and the corresponding answer is a subspan of the corresponding articles.
    \item \textsf{NewsQA}~\cite{trischler2016newsqa}: the documents in NewsQA are articles collected from CNN news. Similar to SQuAD, questions are acquired through crowd-sourcing while the answer is a subspan of documents.
    \item \textsf{DuReader}~\cite{he2018dureader}: DuReader is an open-domain machine reading comprehension dataset in which questions are collected from real anonymized user queries. The documents and the answers are acquired using the search engine. 
\end{itemize}

Besides, the answer should be subspan of the corresponding document. In this setting, we filtered out the data item in DuReader whose answer is not part of the document. 
The QA pair associated with one document should be unique. However, for SQuAD and NewsQA, one long document may have more than one QA pair. For this reason, we split the long document into multiple sub-document to ensure that each sub-document contain only one QA pairs. 
We list the statistics of the modified datasets in Table~\ref{tab:modified_dataset}. 
\begin{table}[H]
  \centering
  \caption{The statistics of the filtered datasets.}
    \resizebox{0.7\columnwidth}{!}{
    \begin{tabular}{lrrr}
    \toprule
    & SQuAD & NewsQA & DuReader \\ 
        \midrule 
    \# Training & 59,819 & 37,688 & 198,532\\
    \# Validation & 1,127 & 1,412 & 1,145\\
    \# Test & 3,000 & 3,000  & 6,000\\
    Avg. len. of document & 27.20 & 29.63 & 144.32\\
    Avg. len. of question & 10.18 & 6.57 & 9.65\\
    Avg. len. of answer & 3.28 & 5.22 & 86.54 \\
    \bottomrule
    \end{tabular}
    }
  \label{tab:modified_dataset}
\end{table}

\subsection{Baselines} \label{sec:experiment:baselines}
To evaluate our proposed OneStop model's performance, we compare our model with two types of baselines. The first type is the models for question generation, which is to evaluate the quality of generated questions. 
The second type is about QA pair generation baselines, which is to evaluate the quality of generated QA pairs. 

\noindent\textbf{Baselines for Question Generation}\quad We used the following models as the baselines for the evaluation of question generation.
\begin{itemize}[leftmargin=*]
    \item \textsf{DeepNQG}: the neural question generation model proposed in \cite{du2017learning}, an end-to-end model implemented with GRU module. 
    \item \textsf{CRF-DeepNQG}: we followed the conventional setting in the D2A2Q approach, which firstly selects the most likely answer span from the document and then utilizes the extracted answer span and the document to generate a question. The answer extraction~(AE) is defined as a sequence tagging task implemented with a BiLSTM-CRF model ~\cite{huang2015bidirectional,du2018harvesting}. The embedding of the document and the extracted answer are concatenated together to generate the answer-specific question. If the answer extraction model predicts more than one answer tag, we randomly selected one span from the span set as the answer to generate the question. If no answer tag is predicted, we viewed the whole document as the selected answer span. 
    \item \textsf{BART-QG}: a fine-tuned model from a pretrained BART~\cite{lewis2020bart} model on the question generation task, whose input is document and output is question. 
    \item \textsf{BART-A2QG}: a fine-tuned model from a pretrained BART~\cite{lewis2020bart} model, whose input is answer and output is question. This model is to explore the utility of answer directly in question generation. 
\end{itemize}

\begin{table*}[!tp]
  \centering
  \caption{The comparison of baselines on question generation.}
    \resizebox{0.96\textwidth}{!}{
    \begin{tabular}{lrrrrrrrrrrrrrrr}
    \toprule
    \multirow{2}{*}{Models} & \multicolumn{5}{c}{SQuAD} & \multicolumn{5}{c}{NewsQA} & \multicolumn{5}{c}{DuReader} \\
    \cmidrule(lr){2-6} \cmidrule(lr){7-11} \cmidrule(lr){12-16}
     & BLEU-1 & BLEU-2 & Rouge-1 & Rouge-2 & Rouge-L & BLEU-1 & BLEU-2 & Rouge-1 & Rouge-2 & Rouge-L & BLEU-1 & BLEU-2 & Rouge-1 & Rouge-2 & Rouge-L \\ 
    \midrule
    {DeepNQG}   & 17.49 & 8.81 & 17.54 & 4.53 & 17.77 & 14.30 & 6.22 & 14.64 & 2.81 & 14.79 & 3.14 & 1.72 & 4.67 & 1.32 & 4.72\\ 
    {CRF-DeepNQG}   & 19.61 & 9.68 & 19.10 & 4.74 & 18.92 & 17.06 & 7.93 & 17.07 & 3.73 & 17.11 & 0.70 & 0.53 & 7.66 & 4.10 & 7.76\\ 
    BART-QG   & \textbf{31.36} & \textbf{21.25} & 32.65 & 14.62 & 29.04 & \textbf{22.30} & \textbf{13.48} & \textbf{23.32} & \textbf{8.34} & 21.81 & 45.22 & 38.33 & \textbf{47.59} & \textbf{33.53} & \textbf{43.13}\\
    BART-A2QG   & 20.85 & 10.51 & 21.50 & 5.50 & 18.81 & 21.53 & 11.81 & 23.29 & 6.96 & 21.75 & 40.61 & 33.94 & 42.85 & 29.32 & 38.58 \\
    \midrule
    \textbf{OneStop}  & \textbf{31.32} & \textbf{21.28} & \textbf{32.77} & \textbf{14.79} & \textbf{29.10} & \textbf{22.28} & \textbf{13.46} & \textbf{23.39} & \textbf{8.39} & \textbf{21.90} & \textbf{45.19} & \textbf{38.35} & \textbf{47.56} & \textbf{33.59} & \textbf{43.16} \\
    \bottomrule
    \end{tabular}
  }
  \label{tab:question_generation}
\end{table*}

\noindent\textbf{Baselines for Question-Answer Pair Generation: \;\;Pipeline vs. OneStop}\quad
For pipeline QA pair generation methods, we utilized the aforementioned question generation model for question generation. To obtain the corresponding answer to the generated question, we chose BERT~\cite{devlin2019bert} as the answer extraction model. 
In this setting, we have the following QA pair generation approaches: 
\begin{itemize} [leftmargin=*]
    \item \textbf{Existing Pipeline Approach}
    \begin{itemize}
        \item \textsf{DeepNQG + BERT-MRC}: the \textsf{DeepNQG} model for question generation and the BERT model for answer extraction.
        \item \textsf{CRF-DeepNQG + BiLSTM-CRF}: as described above, the answer extraction model is implemented with a BiLSTM-CRF model. The BiLSTM-CRF model's tagged phrase is used as the answer $a$ while the answer-aware question is used as the question $q$ corresponding to $a$. 
        \item \textsf{BART-A2QG + BERT-MRC}: the BART-A2QG for question generation and the BERT model for answer extraction. 
        \item \textsf{BART-QG + BERT-MRC}: BART model is used as the question generation model. The BERT model is used for answer extraction. 
    \end{itemize}
    \item \textbf{Our Methods}
    \begin{itemize}
        \item \textsf{OneStop}: our proposed OneStop model involves only one model, in which the process of question generation and answer extraction is simultaneous and affects each other. 
        \item \textsf{OneStop + BERT-MRC}:  the approach in which we used the question generated by OneStop and the answer extracted by the BERT model. 
    \end{itemize}
\end{itemize}

\subsection{Evaluation Metrics}
We evaluated the involved baselines from two aspects: (1) the similarity between generated questions and the ground-truth; (2) the quality of the generated QA pairs.

\noindent\textbf{Similarity Between Generated Questions and Ground-Truth} \quad
We chose BLEU-1, BLEU-2~\cite{papineni2002bleu}, Rouge-1, Rouge-2, and Rouge-L~\cite{lin2004rouge} as the evaluation metrics to evaluate the similarity between the generated questions and the ground-truth questions. 

\noindent\textbf{Quality of Generated Question-Answer pairs} \quad Since there is no widely-accepted automatic metrics on the quality of QA pairs, we used the score from two human annotators as the quality of generated QA pairs. 
The specific scoring criteria of the human raters are given as follows: 
\begin{itemize}[leftmargin=*]
    \item \textsf{Score 0}: if any of the following cases are encountered, the QA pairs is given a score of 0. (1) there are serious grammatical errors in the question; (2) the question is an empty string; (3) the question is totally unrelated to the given document; (4) the question is unanswerable, i.e., the question cannot be answered by referring to the document; (5) The question and the answer are totally unrelated.
    \item \textsf{Score 0.5}: the question is partly related to the document.
    \item \textsf{Score 1}: the question is closely related to the document and is grammatically correct, but the answer is not associated with the question. 
    \item \textsf{Score 1.5}: the question is closely related to the document and is grammatically correct, but the answer can only partially answer the question or contains redundant information. 
    \item \textsf{Score 2}: the question is closely related to the document, and the answer can precisely and concisely reply to the question. 
\end{itemize}
These involved baselines are compared based on an average over human raters' scores. 

\subsection{Model Settings} \label{sec:experiment:settings}
The encoder and the decoder in all the involved pre-trained language models contain 6 layers and a hidden size of 768.
We utilized one well-trained English pre-trained BART model as the initial checkpoint of \textsf{BART-A2QG}, \textsf{BART-QG} and \textsf{OneStop} on SQuAD and NewsQA datasets. 
For models on the DuReader dataset, we pre-trained the BART language model on a very-large Chinese Baike dataset as the initial checkpoint of \textsf{BART-A2QG}, \textsf{BART-QG} and \textsf{OneStop}. 
In our experiments, we set $\gamma=0.2$. The beam size is set to be 3. The batch size is 16 and epoch is set to 4. We chose Adam as our optimizer. The learning rate is set to 1e-4 with a warmup ratio of 0.05. The dropout rate $p=0.1$. All the experiment is run with P100 GPUs. 

\section{Experimental Results and Analysis} \label{sec:result}
\subsection{Results on Question Generation}  \label{sec:result:qg}
We list the baselines' performance on question generation in Table~\ref{tab:question_generation}. As can be observed, \textsf{OneStop} model achieves better or comparable performance than the baselines on question generation.  The performance of DeepNQG is quite poor since the RNN-based model cannot handle the long document well. 
The comparison between \textsf{BART-QG} and \textsf{OneStop} proves that the answer extraction does not degenerate the quality of generated questions. It even improves the performance of generated question models. This phenomenon can be explained by the fact that the probability of answer extraction $p(a \vert d, \hat{q})$ can further enhance the question generation model $p(q \vert d; \theta_{\text{d2q}})$.

\subsection{Results on QA Pair Generation}
We list the result of QA evaluation in Table~\ref{tab:qa_evaluation}. 
From the results, we have the following observations: 
\begin{enumerate}[leftmargin=*]
    \item \textsf{OneStop} outperforms pipeline methods: \textsf{CRF-DeepNQG + BiLSTM-CRF, DeepNQG + BERT-MRC}, and \textsf{BART-A2QG + BERT-MRC} significantly and achieves a human rater's score 1.41, which prove the effectiveness of the OneStop model on question-answer pair generation. 
    \item Compared with \textsf{OneStop}, \textsf{OneStop + BERT-MRC} sees an additional performance improvement.  
    The difference between \textsf{OneStop}~(1.41) and \textsf{OneStop + BERT-MRC}~(1.67) proves that the answer extraction module in OneStop model is not as good as the answer extraction model implemented with BERT. This can be explained by the fact that the BERT model for answer extraction has 12 transformer layers, which has better representation capacity on the document and question encoding. However, both the encoder and decoder in the OneStop model have only 6 transformer layers, which has a less satisfying representation ability. 
    \item The comparison between \textsf{BART-QG + BERT-MRC} and \textsf{OneStop + BERT-MRC} is to verify the effectiveness of answer extraction module on OneStop's question generation module. The improvement~(from 1.61 to 1.67) demonstrates that the answer extraction module in the OneStop model enhances the quality of QA pairs. 
\end{enumerate}
\begin{table}[ht!]
  \centering
  \caption{Result of generated question-answer pairs.}
    \resizebox{0.99\columnwidth}{!}{
    \begin{tabular}{lr|lr}
    \toprule
    Approach & Score & Approach & Score \\ 
    \midrule 
    CRF-DeepNQG + BiLSTM-CRF & 0.22  & BART-QG + BERT-MRC & 1.61\\
    DeepNQG + BERT-MRC & 0.20 & OneStop & 1.41 \\
    BART-A2QG + BERT-MRC & 0.24 & OneStop + BERT-MRC & \textbf{1.67}\\
    \bottomrule
    \end{tabular}
    }
  \label{tab:qa_evaluation}
\end{table}

\subsection{Model Efficiency}
Another significant advantage of our OneStop model is efficiency:
\begin{enumerate} [leftmargin=*]
    \item We list the number of parameters of each QA pair generation approach in Table~\ref{tab:efficiency}. As we can see, OneStop model is one of the lightest model for QA pair generation. 
    \item Different from the pipeline approaches that involve more than one model, which requires additional efforts and computational resource to train and deploy these models. OneStop, nevertheless, involves only one model, which is much more efficient for both training and deployment. \item The pipeline baselines require additional human efforts when deployed online. For instance, for the answer extraction~(AE) model in D2A2Q, it may select none or more than one answer span, which requires well-designed rules to select from these answer span for question generation. 
\end{enumerate}

\begin{table}[ht!]
  \centering
  \caption{The number of parameters~(millions) of each QA pair generation approach.}
    \resizebox{0.7\columnwidth}{!}{
    \begin{tabular}{lrrr}
    \toprule
    Approach & SQuAD & NewsQA & DuReader \\ 
    \midrule 
    CRF-DeepNQG + BiLSTM-CRF & 109 & 71 & 163 \\
    DeepNQG + BERT-MRC & 151 & 146 & 323\\
    BART-A2QG + BERT-MRC & 248 & 248 & 423\\
    BART-QG + BERT-MRC & 248 & 248 & 423 \\
    OneStop & {142} & {142} & {121} \\
    OneStop + BERT-MRC & {253} & {253} & {427} \\

    \bottomrule
    \end{tabular}
    }
  \label{tab:efficiency}
\end{table}

\subsection{Case Study}
We present several QA pairs generated by our OneStop model in Table~\ref{tab:case_study}. We also include the answer span predicted by the BERT model. Based on our observation, in most cases, the answer predicted by the OneStop model is same as that of BERT. But for certain cases, the answer extraction in OneStop may tend to include the related information besides the precise answer span. Most of the extracted QA pairs from our OneStop model can be applied in downstream tasks like question answering. 
\begin{table}[ht!]
    \centering
    \caption{Question-answer pairs generated by OneStop.}
     \resizebox{0.85\columnwidth}{!}{
    \begin{tabular}{p{1.5cm}p{8.2cm}}
    \toprule
     & text \\ \midrule  
    \textit{document} & The French crown's refusal to \hlyellow{allow non-Catholics to settle in New France} may help to explain that colony's slow rate of population growth compared to that of the neighbouring British colonies, which opened settlement to religious dissenters. \\ \cmidrule(lr){2-2} 
    \textit{OneStop question} & What did the French government refusal to allow?  \\  \cmidrule(lr){2-2}
    \textit{BERT-MRC answer} & non-Catholics \\ \cmidrule(lr){2-2}
    \textit{OneStop \quad answer} & non-Catholics to settle in New France \\ \midrule
     \textit{document} & The delta is delimited in the West by the \hlcyan{Alter Rhein ("Old Rhine")} and in the East by a modern canalized section. \\ \cmidrule(lr){2-2}
    \textit{OneStop question} & What is the delta delimited by? \\  \cmidrule(lr){2-2}
    \textit{BERT-MRC answer} & Alter Rhein \\ \cmidrule(lr){2-2}
    \textit{OneStop \quad answer} & Old Rhine \\
    \bottomrule
    \end{tabular}
    }
    \label{tab:case_study}
\end{table}
\section{Conclusion} \label{sec:conclusions}
Existing pipeline QA pair generation approaches suffer problems like incompatible and sub-optimal solutions, inefficiency, and heavy human effort.  
This paper proposes a transformer-based sequence-to-sequence model to generate QA pairs in a one-stop fashion. Our model achieves state-of-the-art performance on question generation and QA pair generation on three large-scale machine reading comprehension datasets in a more efficient way. Our work sheds light on a novel One-Stop approach to QA pair extraction. We will explore more effective techniques of generating QA pairs such as the soft approach of answer extraction and copy mechanism.

\bibliographystyle{ACM-Reference-Format}
\bibliography{sample-base}

\end{document}